\documentclass[review]{elsarticle}
\makeatletter
\def\ps@pprintTitle{%
 \let\@oddhead\@empty
 \let\@evenhead\@empty
 \def\@oddfoot{}%
 \let\@evenfoot\@oddfoot}
\makeatother
\usepackage{tikz}
\usepackage{comment}
\usepackage{amsmath,amssymb} 
\usepackage{color}
\usepackage{epsfig}
\usepackage[inline]{enumitem}
\usepackage[caption=false]{subfig}
\usepackage{soul}
\usepackage[utf8]{inputenc} 
\usepackage[T1]{fontenc}    
\usepackage{booktabs}       
\usepackage{amsfonts}       
\usepackage{nicefrac}       
\usepackage{microtype}      
\usepackage{algorithmic}
\usepackage{algorithm}
\usepackage{numprint}
\usepackage{siunitx}
\usepackage{etoolbox}
\newcommand{\ubold}{\fontseries{b}\selectfont}
\robustify\ubold
\usepackage{bbold}
\usepackage{wrapfig}

\usepackage[caption=false]{subfig}
\usepackage{pifont}

\usepackage{lineno,hyperref}
\usepackage{array,multirow,graphicx}

\makeatletter
\newcommand\footnoteref[1]{\protected@xdef\@thefnmark{\ref{#1}}\@footnotemark}
\makeatother

\modulolinenumbers[5]

\journal{Journal of Pattern Recognition}

\begin{document}

\pagenumbering{gobble}

\let\WriteBookmarks\relax
\def\floatpagepagefraction{1}
\def\textpagefraction{.001}
\pagenumbering{arabic}

\newcommand{\ttinneriter}{\texttt{inner\_iters}}
\newcommand{\ttit}{\texttt{iter}}
\newcommand{\ttmaxiter}{\texttt{max\_iter}}
\newcommand{\ttbestScore}{\texttt{best\_score}}
\newcommand{\ttbestR}{\texttt{best\_rotation}}
\newcommand{\ttbestT}{\texttt{best\_translation}}

\newcommand{\cL}{\mathcal{L}}
\newcommand{\cM}{\mathcal{M}}
\newcommand{\cN}{\mathcal{N}}
\newcommand{\cI}{\mathcal{I}}
\newcommand{\cS}{\mathcal{S}}
\newcommand{\cD}{\mathcal{D}}
\newcommand{\cP}{\mathcal{P}}
\newcommand{\cQ}{\mathcal{Q}}
\newcommand{\cO}{\mathcal{O}}
\newcommand{\cT}{\mathcal{T}}
\newcommand{\cad}{\mathcal{d}}
\newcommand{\cX}{\mathcal{X}}
\newcommand{\cXh}{\hat{\mathcal{X}}}
\newcommand{\cC}{\mathcal{C}}
\newcommand{\cF}{\mathcal{F}}

\newcommand{\be}{\mathbf{e}}
\newcommand{\br}{\mathbf{r}}
\newcommand{\bx}{\mathbf{x}}
\newcommand{\bxh}{\hat{\mathbf{x}}}
\newcommand{\bX}{\mathbf{X}}
\newcommand{\bY}{\mathbf{Y}}
\newcommand{\bZero}{\mathbf{0}}
\newcommand{\hbX}{\hat{\mathbf{X}}}
\newcommand{\bS}{\mathbf{S}}
\newcommand{\bs}{\mathbf{s}}
\newcommand{\bp}{\mathbf{p}}
\newcommand{\bq}{\mathbf{q}}
\newcommand{\bD}{\mathbf{D}}
\newcommand{\bd}{\mathbf{d}}
\newcommand{\bA}{\mathbf{A}}
\newcommand{\bR}{\mathbf{R}}
\newcommand{\bt}{\mathbf{t}}
\newcommand{\bH}{\mathbf{H}}
\newcommand{\bh}{\mathbf{h}}
\newcommand{\by}{\mathbf{y}}
\newcommand{\bz}{\mathbf{z}}
\newcommand{\bu}{\mathbf{u}}
\newcommand{\ba}{\mathbf{a}}
\newcommand{\bg}{\mathbf{g}}
\newcommand{\bo}{\mathbf{o}}
\newcommand{\bl}{\mathbf{l}}
\newcommand{\bOnes}{\mathbf{1}}
\newcommand{\bF}{\mathbf{F}}
\newcommand{\bK}{\mathbf{K}}
\newcommand{\bI}{\mathbf{I}}
\newcommand{\tdf}{\tilde{f}}
\newcommand{\tdh}{\tilde{h}}

\newcommand{\bbR}{\mathbb{R}}
\newcommand{\bbE}{\mathbb{E}}
\newcommand{\bbD}{\mathbb{D}}
\newcommand{\bbF}{\mathbb{F}}
\newcommand{\bbFh}{\hat{\mathbb{F}}}
\newcommand{\bmu}{\boldsymbol{\mu}}
\newcommand{\bhr}{\hat{\mathbf{r}}}
\newcommand{\bJ}{\mathbf{J}}
\newcommand{\Nsample}{$N_\text{sample}$}

\newcommand{\kernel}{\psi}

\newcommand{\residual}{\mathbf{r}}
\newcommand{\btheta}{\boldsymbol{\theta}}

\newcommand{\cV}{\mathcal{V}}
\newcommand{\cE}{\mathcal{E}}
\newcommand{\cB}{\mathcal{B}}
\newcommand{\cG}{\mathcal{G}}

\newcommand{\xmark}{\ding{55}}%
\newcommand{\cmark}{\ding{51}}%

\begin{frontmatter}

\title{Multi-Camera Multi-Object Tracking on the Move via \\ Single-Stage Global Association Approach}

\author[2]{Pha {Nguyen}\footnote{\label{note}equal contribution}}

\author[1]{Kha Gia {Quach}\footnotemark} 

\author[1]{Chi Nhan {Duong}}

\author[3]{Son Lam {Phung}}

\author[2]{Ngan {Le}}

\author[2]{Khoa {Luu}}

\address[2]{Computer Science and Computer Engineering Department, University of Arkansas, Fayetteville, USA}
\address[1]{Computer Science and Software Engineering Department, Concordia University, Montreal, QC, CANADA}
\address[3]{Faculty of Engineering and Information Sciences, University of Wollongong, NSW, Australia}

\begin{abstract}

The development of autonomous vehicles generates a tremendous demand for a low-cost solution with a complete set of camera sensors capturing the environment around the car. It is essential for object detection and tracking to address these new challenges in multi-camera settings. In order to address these challenges, this work introduces novel Single-Stage Global Association Tracking approaches to associate one or more detection from multi-cameras with tracked objects. These approaches aim to solve fragment-tracking issues caused by inconsistent 3D object detection. Moreover, our models also improve the detection accuracy of the standard vision-based 3D object detectors in the nuScenes detection challenge. The experimental results on the nuScenes dataset demonstrate the benefits of the proposed method by outperforming prior vision-based tracking methods in multi-camera settings.

\end{abstract}

\end{frontmatter}

\section{Introduction}

Object detection and tracking have become two of the most critical tasks in autonomous vehicles (AV). Recent developments in deep learning methods have dramatically boosted the performance of object understanding and tracking in autonomous driving applications.

\begin{figure}[t]
  \centering
  \includegraphics[width=0.9 \linewidth]{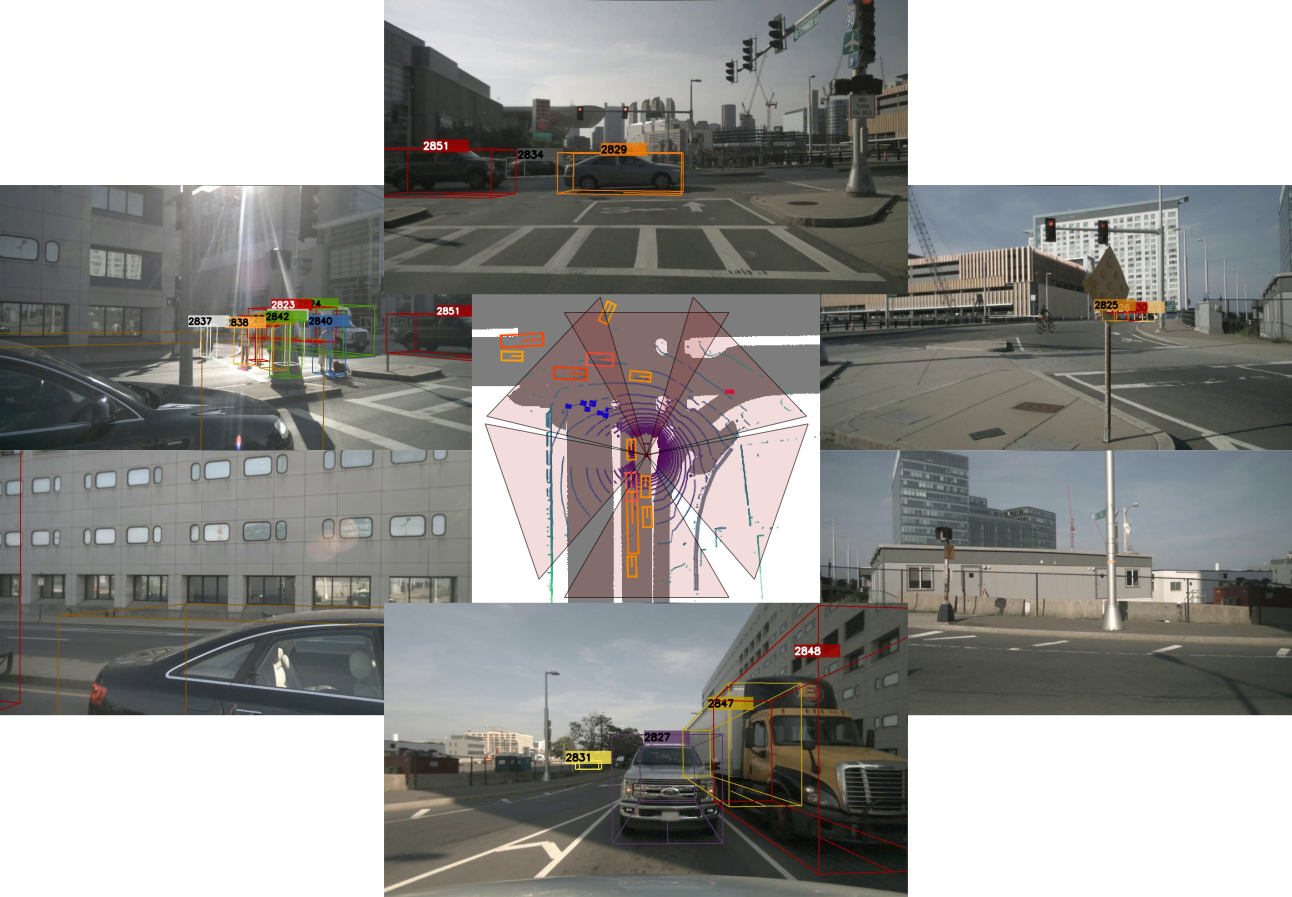}
  \caption{Sample of multi-view captured via a multi-camera setup on a vehicle from nuScenes \cite{caesar2020nuscenes}.}\label{fig:sample_nuscenes_a}
\end{figure}

\begin{figure}[t]
  \centering
  \includegraphics[width=0.9 \linewidth]{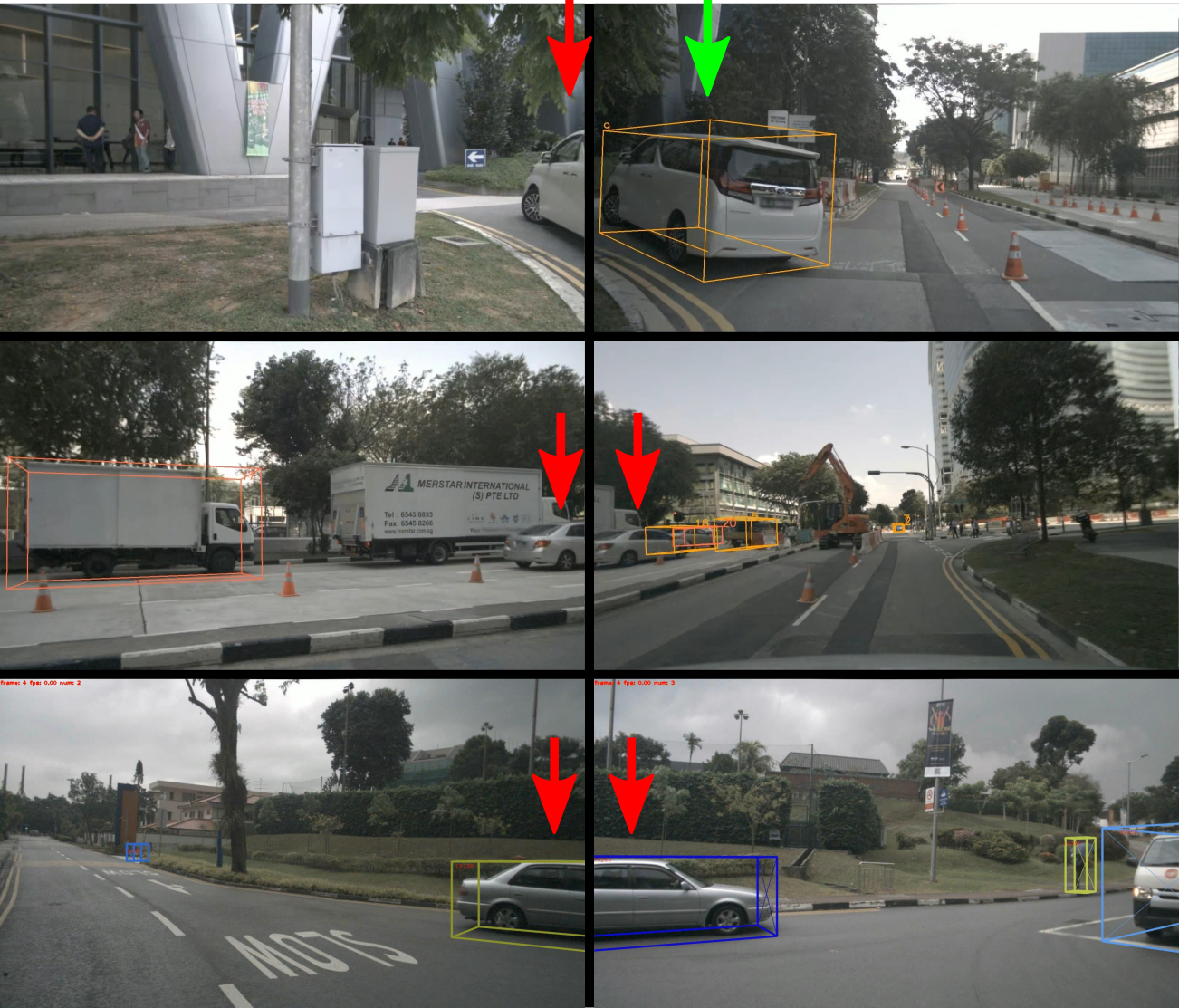}
  \caption{First row: the object detector KM3D \cite{2009.00764} fails to detect partial objects in one camera but can detect them in another. Second row: The detector fails to detect objects in both cameras. Third row: the SC-MOT method DEFT \cite{Chaabane2021deft} fragments a global object ID into many local IDs when it moves across cameras. The green arrow indicates the true-positive detection sample; the red arrows indicate false-negative detection and tracking samples.}\label{fig:sample_nuscenes_b}
\end{figure}

The object tracking problem in AVs is far apart from multiple camera multiple object tracking (MC-MOT) \cite{quach2021dyglip} in surveillance settings where cameras are stationary, i.e., their positions are fixed, but their poses may change in PTZ cameras cases. For clarity, MC-MOT in surveillance settings is referred to as \textit{static MC-MOT} and MC-MOT in AVs as \textit{dynamic MC-MOT on-the-move} since cameras are moving with the vehicle. 
Other works \cite{vo2020self} consider tracking the activities of people from multiple moving cameras, where the movements are subtle with large overlapping regions between cameras. In contrast, our camera setting in this paper contains large movements and small overlapping regions between cameras as the traveling car passes by other objects.
Such a setup with some redundancy, i.e., certain overlapping fields-of-view, presents some new challenges for MOT to work with 3D object detectors to track objects and maintain the stability of predictions across video frames in multiple views.

As a result, camera-based tracking methods in the current leaderboard of autonomous driving datasets, e.g., nuScenes \cite{caesar2020nuscenes} and Waymo \cite{sun2019scalability}, appear to be using only single-camera settings. However, the datasets were collected in multi-camera settings as shown in \ref{fig:sample_nuscenes_a}. Thus, this work aims to use redundant data to improve detection and tracking performance.

In the \textit{MC-MOT} settings, traditional two-stage approaches \cite{cai2014exploring, chen2016equalized, quach2021dyglip, nguyen2022multi} track objects on each camera independently, i.e., single-camera tracking (SCT), then link-local tracklets across cameras via global matching steps based on Re-ID features. Applying such a two-stage approach to \textit{dynamic MC-MOT} settings on AVs leads to a problem with the global matching that relies on complicated graph structures to assign a global ID to all detection.
In addition, this approach cannot handle scenarios when the detector fails to detect objects from one of the cameras \cite{Pha2022icip, Le2016cvprw, Le2016icpr, Yutong2016}. Moreover, it requires additional steps to merge many local IDs, as shown in Fig. \ref{fig:sample_nuscenes_b}.
Therefore, there are better solutions than using SCT multiple times.

\subsection{Contributions of this Work}

This work presents a single-stage MC-MOT approach directly using the outputs of an object detector as the inputs instead of SCT trajectories.
To achieve this goal, we mathematically reformulate association steps in static MC-MOT into a single global association step as a \textit{one-to-many assignment problem} to match one target, i.e., tracked objects in the world coordinate, with multiple detection, i.e., objects appear in multi-camera overlapping regions.
This assignment can be solved efficiently via our proposed \textit{Fractional Optimal Transport Assignment (FOTA}) method.
Moreover, since this assignment problem can be defined in both the traditional track-by-detection scheme and the more recent track-by-attention scheme, we demonstrate its ability in both our proposed \textit{Single-Stage Global Assignment (SAGA)} schemes as \textit{SAGA-Track} and \textit{SAGA-TrackNet}, respectively.
Evaluate proposed methods with a comprehensive evaluation criterion to demonstrate their robustness compared to previous frameworks. The proposed method reduces the IDSwitch error from 3,807 to 870 and improves the tracking performance by up to 6.4\% on the nuScenes Test Set benchmark.

\section{Related Work}

The MOT problem on AVs has recently received much attention in the research community. Recent methods in static MC-MOT settings have been reviewed in \cite{quach2021dyglip}, while dynamic MC-MOT settings are still an open research area. The most recent work reviewed in this section is focused on the assignment or association formulation in SC-MOT and static MC-MOT.

\textbf{Assignment in SC-MOT.}
While many works \cite{Bewley2016_sort, Wojke2017simple} calculated the assignment costs between tracklets and detection by using some distance measurements over deep features or locations, some approaches directly computed the similarity scores. Xiang et al. \cite{7410891} built a bipartite graph over the affinity computed by the LSTM as edge cost and solved the association by the Hungarian algorithm. Ran et al. \cite{10.1007/978-3-030-05710-7_34} proposed a Pose-based Triple Stream Network to extract three kinds of similarity scores, i.e., appearance, motion, and interaction, and then fuse the average strategy into a final similarity score in a bipartite graph by the greedy match algorithm.

\textbf{Assignment in Static MC-MOT.} He et al. \cite{he2020multi} constructed a global similarity matrix from local tracklets in all single views and then estimated targets' trajectory by offline performing Matrix Factorization. Ristani and Tomasi \cite{8578730} solved the ID assignment task by correlation clustering, then executed interpolation and elimination to fill the gap and filter indecisive tracks. 
Quach et al. \cite{quach2021dyglip} proposed a dynamic graph to transform pre-computed Re-ID features into new context-aware ID features. Hence it performs better clustering and yields more accurate results. Yoon et al. \cite{8387029} maintained a set of track hypotheses all the time by the Multiple Hypothesis Tracking algorithms and also reduced the excess by introducing a gating mechanism for tree pruning. Zhang et al. \cite{Zhang2017MultiTargetMT} utilized the Re-Ranking algorithm \cite{8099872} on the global cost matrix to cluster IDs. However, directly applying these approaches to the \textit{dynamic} setting on AVs suffers from a significant real-time performance decrease, computation complexity, and domain irrelevance. Therefore, several methods to solve object tracking on the fly have been proposed, as referred to in the following parts.

\textbf{Using Motion Models.}
Weng et al. \cite{weng2020ab3dmot} proposed a simple yet effective baseline to utilize a classic state estimator (the Kalman Filter) for tracking 3D bounding boxes. These bounding boxes can be obtained from a \mbox{LiDAR} point cloud object detector \cite{Shi_2019_CVPR, 2019arXiv190809492Z, qi2016pointnet, qi2017pointnetplusplus, zhou2017voxelnet} or an image-based object detector \cite{Ren17CVPR, zhou2019objects, Simonelli_2019_ICCV, Hu3DT19}. Chiu et al. \cite{chiu2020probabilistic} improved the Kalman Filter tracking system using the Mahalanobis distance between the predicted states and observations. The method is reasonably effective in filtering outliers and handling partially and fully occluded objects.

\textbf{Using Appearance Models.}
Zhou et al.'s approaches \cite{zhou2019objects, zhou2020tracking} are widely used for single-camera tracking. These approaches simplify the tracking procedure by treating objects as points, which usually involves many computationally intensive steps from detection to assigning object ID.
Hu et al. \cite{Hu3DT19} estimated robust 3D box information from 2D images and adopted 3D box-reordering and LSTM as a motion module to link objects across frames.

\textbf{Using Hybrid Approaches.}
Chaabane et al. \cite{Chaabane2021deft} trained the object detection and the object association task simultaneously by adding a feature extractor and a matching head after the object detector. In addition, an LSTM instead of a Kalman Filter is used for motion prediction. Yin et al. \cite{yin2021center} followed a similar process but performed feature extraction on point cloud maps.

\textbf{Using Modern Approaches.}
Graph Neural Network \cite{weng2020gnn3dmot}, Self-Attention \cite{zhu2018online}, and Transformer \cite{vaswani2017attention} have led to a new learning-from-context paradigm. This paradigm has attracted considerable research attention recently because of its promising performance in a wide range of tasks from natural language processing \cite{ott2018scaling, devlin2019bert, Radford2018ImprovingLU, liu2019roberta} to computer vision \cite{dosovitskiy2020, carion2020endtoend, wang2020endtoend, ramachandran2019standalone, touvron2021training, zhu2020deformable}. A limited number of these methods have been applied for \textit{dynamic} MC-MOT in autonomous vehicles \cite{nguyen2022multi}, apart from many SC-MOT approaches \cite{Gao_2019_CVPR, Chu_2017_ICCV, sun2020transtrack, meinhardt2021trackformer, Zhu_2018_ECCV, Weng2020_GNN3DMOT, Weng2020_GNNTrkForecast}. Weng et al. \cite{Weng2020_GNN3DMOT} proposed the first feature interaction method that leverages a Graph Neural Network to adapt features from one object to another individual. Meinhardt et al. \cite{meinhardt2021trackformer} proposed a new tracking-by-attention paradigm (compared to the existing tracking-by-regression, tracking-by-detection, and tracking-by-segmentation) to deal with occlusions and determine the tracker's Spatio-temporal correspondences. Sun et al. \cite{Zhu_2018_ECCV} utilized the Query-Key mechanism to perform joint detection-and-tracking and disentangle complex components in previous tracking systems.

Compared to these prior works, the critical difference in our approach is
that it uses a world coordinate system in AVs multi-camera system to solve the global one-to-many association step. It is possible by matching one tracked object with multiple detection. Thus, it eliminates the need for another association step, i.e., using Re-ID, and reduces the effort of adopting several empirical rules and heuristics to handle overlapping FOVs.

\section{Our Proposed Method}

This section presents the proposed dynamic MC-MOT approaches with a one-to-many global assignment method.

\subsection{Problem Definition}
\label{sec:problem_def}

Given video frames from $K$ cameras at the $t$-th time step, denoted by the set ${\mathcal{I}^{(t)}=\{I_1^{(t)},\dots, I_k^{(t)}\dots,I_K^{(t)}\}}$, MC-MOT system provides a set of detected objects $ \mathcal{O}^{(t)} = \{ \mathbf{o}_{j}^{(t)}\}$ associated with their identities. Object bounding boxes and classes can be predicted using an object detector given each frame in  $\mathcal{I}^{(t)}$ separately.
The identities of objects are obtained by associating with tracklets, i.e., a set of bounding boxes 
with a track ID $i$ as $\cT_i = \{ \mathbf{tr}^{(t_1)}_{i}, \mathbf{tr}^{(t_2)}_{i}, \cdots \}$.
Objects detected on each camera and track are represented by 3D bounding boxes in world coordinates. Note that tracklets are shared across cameras and are often referred to as a global track ID. During $T$ frames of a video sequence, the sub-sequence of $(t_1, t_2, \cdots)$ is the time steps when the tracked object appears within the camera views. Each track $ \mathbf{tr}^{(t)}_{i} $ is estimated using a motion model from the previous frame $t-1$ and then updated with the detection of the corresponding tracked objects as follows,
\begin{equation}
  \footnotesize
  \begin{split}
    \hat{\mathbf{tr}}^{(t)}_{i} &= \mathcal{M}_{\text{pred}} (\mathbf{tr}^{(t - 1)}_{i}) \\
    \mathbf{tr}^{(t)}_{i} &= \mathcal{M}_{\text{update}} (\hat{\mathbf{tr}}^{(t)}_{i}, \mathbf{o}^{(t)}[i])
  \end{split}
\end{equation}
\begin{equation}
  \footnotesize
  \text{where } \mathbf{o}^{(t)} [i] =
  \begin{cases}
    \mathbf{o}^{(t)}_{j} & \text{if detected object } \mathbf{o}^{(t)}_{j} \text{ associates with the $i$-th tracklet}\\
    \varnothing            & \text{if no object } \text{ associates with the $i$-th tracklet}
  \end{cases}
\end{equation}
\noindent
Here, $\mathcal{M}_{\text{pred}}$ is a function or a network to predict the following location of the track based on the motion model, and $\mathcal{M}_{\text{update}}$ is a function to update the location of the track in the current time step $t$. In this paper, we use two different motion models, i.e., linear Kalman Filter and Non-linear Transfomer-based Network.
To know which detected object $j$ is being used to update the corresponding tracklet $i$. Each detection is then assigned to a tracklet based on a matching algorithm with a cost function. It also determines whether the detection is a new or existing object from the previous frame.
Generally, the cost functions to match detection with tracklets can be defined as in Eqn. \eqref{eq:cijk}.
\begin{equation} \label{eq:cijk}
  \footnotesize
  c_{ij} = \cC_{\text{match}} [i, j] =d\left(\hat{\mathbf{tr}}^{(t)}_{i} , \mathbf{o}_{j}^{(t)} \right)
\end{equation}
where $d(\cdot, \cdot)$ is the distance between the detected and tracked objects.
Several distance metrics can be adopted for $d(\cdot, \cdot)$ such as Mahalanobis distance \cite{mahalanobis1936generalized} implemented in \cite{chiu2020probabilistic}, 2D or 3D GIoU \cite{rezatofighi2019generalized}.

\begin{equation}
\label{eq:mahalanobis}
\footnotesize  
    d_{\text{Mahalanobis}} \left(\hat{\mathbf{tr}}^{(t)}_{i} , \mathbf{o}_{j}^{(t)} \right) = \sqrt{(\mathbf{o}_{j}^{(t)} - \hat{\mathbf{tr}}^{(t)}_{i})^{T} {\mathbf{S}^{(t)}}^{-1} (\mathbf{o}_{j}^{(t)} - \hat{\mathbf{tr}}^{(t)}_{i})}, 
\end{equation}
\begin{equation}
  \label{eq:g_iou}
  \footnotesize
    d_{\text{GIoU}} \left(\hat{\mathbf{tr}}^{(t)}_{i} , \mathbf{o}_{j}^{(t)} \right) = 1 - \left(\frac{|\mathbf{o}_{j}^{(t)} \cap \hat{\mathbf{tr}}^{(t)}_{i}|}{|\mathbf{o}_{j}^{(t)} \cup \hat{\mathbf{tr}}^{(t)}_{i}|} - \frac{|\mathbf{cv}^{(t)}_{ij} \setminus (\mathbf{o}_{j}^{(t)} \cup \hat{\mathbf{tr}}^{(t)}_{i})|}{|\mathbf{cv}^{(t)}_{ij}|}\right),
\end{equation}

Here, $\mathbf{S}^{(t)}$ is the covariance that represents the uncertainty of the predicted object state as implemented in \cite{chiu2020probabilistic}. In addition, $\mathbf{cv}^{(t)}_{ij}$ is the smallest enclosing convex shape of $\mathbf{o}_{j}^{(t)}$ and $\hat{\mathbf{tr}}^{(t)}_{i}$.
Note that multi-view geometry is implicitly applied when we compute the distance metrics above.

\subsection{Single-Stage Global Assignment Tracking Approach (SAGA-Track)}
\label{sec:bipartite_matching}

With the cost matrix defined above, the assignment algorithm has to assign the detected objects to the correct tracklets. To assign detection to tracklets, a straightforward approach is to pool all detection and tracks into two corresponding sets and perform an \textit{one-to-one matching} algorithm, i.e., Hungarian algorithm, based on a cost matrix similar to the SC-MOT case. However, in dynamic MC-MOT settings, one object can appear in several cameras simultaneously due to camera overlapping.
That means two or more detected objects $\mathbf{o}^{(t)}_{j}$ in different cameras should be matched to one tracklet $\hat{\mathbf{tr}}^{(t)}_{i}$ only.
Therefore, the \textit{one-to-one matching} algorithm cannot handle detection from multiple cameras put together as only one instance of an object in a camera can be matched to the target tracklet causing the remaining detection of that object in other cameras to be unmatched. These unmatched instances may create new tracklets during the tracking process, and a second association step is needed to connect them. It is referred to as the global baseline association in our experiments.

To further equip a tracking system with the capability of tracking multiple instances of the same objects in different cameras, we propose to cast this assignment process to a distribution matching task where tracklets and all detected objects at $t$-th time step can be formed into two distributions.
Formally, let $\mathcal{X} = \{ \hat{\mathbf{tr}}^{(t)}_{i}\}_{i=1}^N$ and $\mathcal{Y} = \{ \mathbf{o}^{(t)}_j\}_{j=1}^M$ be the sets of \textit{$N$ current tracklets} and $M$ \textit{detected objects} from all $K$ cameras at the $t$-th time step.
Let $\mathbf{p}$ and $\mathbf{q}$ be the empirical distributions defined over $\mathcal{X}$ and $\mathcal{Y}$, respectively. The set of all possible couplings $\Pi (\mathbf{p}, \mathbf{q})$ to transport the mass, i.e., number of object entities, from $\mathcal{X}$ to $\mathcal{Y}$ is defined as in Eqn. \eqref{eqn:TrackletObjCoupling}.
\begin{equation} \label{eqn:TrackletObjCoupling}
  \footnotesize
    \Pi (\mathbf{p}, \mathbf{q}) =
      \begin{cases}   & \boldsymbol{\pi} \in \mathbb{R}_+^{|\mathbf{p}| \times |\mathbf{q}|}:
               \\ & \boldsymbol{\pi}\mathbb{1}_{|\mathbf{q}|} \leq \mathbf{p}, \boldsymbol{\pi}^{\top}\mathbb{1}_{|\mathbf{p}|} \leq \mathbf{q},\mathbb{1}_{|\mathbf{p}|}^{\top} \boldsymbol{\pi} \mathbb{1}_{|\mathbf{q}|} = s
      \end{cases} \Bigg\}
\end{equation}
where $\pi_{ij}$ denotes the amount of a mass $p_i$ at $\hat{\mathbf{tr}}^{(t)}_{i}$ being associated with the mass $q_j$ at $\mathbf{o}^{(t)}_j$. The inequality in Eqn. \eqref{eqn:TrackletObjCoupling} indicates the possibility of fractional entities being matched between the two distributions as (1) one tracklet in $\mathcal{X}$ can associate with no detection (i.e., the tracked object does not appear in all cameras) or many detections (i.e., the tracked object appears in many cameras); and (2) one detection in $\mathcal{Y}$ can be assigned to zero or one tracklet in $\mathcal{X}$. Moreover, different from the standard Optimal Transport (OT) based approach \cite{villani2009optimal} where the two distributions are required to have the same total probability mass, i.e., $||\mathbf{p}||_1 = ||\mathbf{q}||_1$, and all the mass has to be transported. Eqn. \eqref{eqn:TrackletObjCoupling} \textit{\textbf{focuses on transporting only a fraction $s$ of the mass between two distributions.}} Thus, we named this approach as \textbf{Fractional OT Assignment (FOTA).}

Let $\mathbf{C}=(c_{i,j})$ be the transportation cost matrix where $c_{i,j}$ measures a cost to associate from $\hat{\mathbf{tr}}^{(t)}_{i}$ to $\mathbf{o}^{(t)}_j$.
The proposed FOTA addresses the problem of finding the best assignment solution $\pi$ that minimizes the transportation cost between two distributions:
\begin{equation}
  \label{eq:optimal_transport}
  \footnotesize
  \min_{\boldsymbol{\pi} \in \Pi(\mathbf{p}, \mathbf{q})} \langle \mathbf{C}, \boldsymbol{\pi}\rangle_F = \min_{\boldsymbol{\pi} \in \Pi(\mathbf{p}, \mathbf{q})} \sum_{i}^N \sum_{j}^M c_{ij}\pi_{ij}
\end{equation}
To address the constraints of only transporting a fraction $s$ of mass in Eqn. \eqref{eqn:TrackletObjCoupling}, we propose attaching one more row and column in the cost matrix to handle the mass difference between two distributions as in Eqn. \eqref{eq:extend_cost}.
\begin{equation}
  \label{eq:extend_cost}
  \footnotesize
  \bar{\mathbf{C}} =
  \begin{bmatrix}
    \mathbf{C}                                   & \mathcal{E} \mathbb{1}_{|\mathbf{q}|} \\
    \mathcal{E} \mathbb{1}_{|\mathbf{p}|}^{\top} & 2\mathcal{E} + \max(\mathbf{C})       \\
  \end{bmatrix}
\end{equation}
where $\mathcal{E}$ is a scalar for the bound. If we set the mass of the additional track and object as $p_{N+1} = \| \mathbf{q} \|_1 - s $ and $q_{M+1} = \| \mathbf{p} \|_1 - s $, finding the best assignment solution $\pi$ can be reduced to an unconstrained problem $ \min_{\bar{\boldsymbol{\pi}} \in \Pi(\bar{\mathbf{p}}, \bar{\mathbf{q}})} \langle \bar{\mathbf{C}}, \bar{ \boldsymbol{\pi}} \rangle_F $, where $\bar{\mathbf{p}} = [\mathbf{p}, \| \mathbf{q} \|_1 - s]$ and $ {\bar{\mathbf{q}} = [\mathbf{q}, \| \mathbf{p} \|_1 - s]} $.

\textbf{Solving the One-to-many Assignment.} From the above formula for the Optimal Transport-based Assignment problem in Eqn. \eqref{eq:optimal_transport}, one can solve it in polynomial time as it is a linear program.
However, when there are multiple detected objects and tracklets, the resulting linear program can be large. This issue can be addressed by a fast iterative solution named Sinkhorn-Knopp \cite{cuturi2013sinkhorn}, which converts the optimization target in Eqn. \eqref{eq:optimal_transport} into a non-linear but convex form by adding a regularization term $E$ as in Eqn. \eqref{eq:ot_reg}.
\begin{equation}
  \label{eq:ot_reg}
  \footnotesize
  \min_{\bar{\boldsymbol{\pi}} \in \Pi(\bar{\mathbf{p}}, \bar{\mathbf{q}})} \sum_{i}^N \sum_{j}^M c_{ij} \pi_{ij} + \gamma E\left( \bar{\pi}_{ij} \right)
\end{equation}
where $E( \bar{\pi}_{ij} ) = \bar{\pi}_{ij} ( \log(\bar{\pi}_{ij}) - 1)$. Here, $\gamma$ is a constant regularization term. The constraint optimization target in Eqn. \eqref{eq:ot_reg} can be converted to a non-constraint target using the Lagrange Multiplier method as in Eqn. \eqref{eq:Lagrange}.
\begin{eqnarray} \label{eq:Lagrange}
  \footnotesize
  \min_{\boldsymbol{\bar{\pi}} \in \Pi(\bar{\mathbf{p}}, \bar{\mathbf{q}})} \sum_{i}^N \sum_{j}^M c_{ij} \bar{\pi}_{ij} + \gamma E\left( \bar{\pi}_{ij} \right) + \alpha_j \left( \boldsymbol{\bar{\pi}}^{\top}\mathbb{1}_{|\bar{\mathbf{p}}|} - \bar{\mathbf{q}} \right) + \beta_i \left(  \boldsymbol{\bar{\pi}}\mathbb{1}_{|\bar{\mathbf{q}}|} - \bar{\mathbf{p}} \right)
\end{eqnarray}
where $\alpha_j(j = 1,2,...M)$ and $\beta_i(i = 1,2,...,N)$ are Lagrange multipliers. By letting the derivatives of the optimization target equal 0, the optimal plan $\boldsymbol{\bar{\pi}}^{\star}$ is resolved as:
\begin{equation}
  \footnotesize
  \bar{\pi}^{\star}_{ij} = \exp \left( -\frac{\alpha_j}{\gamma} \right) \exp \left(-\frac{c_{ij}}{\gamma} \right) \exp \left( -\frac{\beta_i}{\gamma} \right)
\end{equation}
Let $u_j = \exp \left( -\frac{\alpha_j}{\gamma} \right), v_i = \exp \left( -\frac{\beta_i}{\gamma} \right), \mathbf{W}[i,j] = \exp \left(-\frac{c[i,j]}{\gamma} \right)$, the following constraints can be enforced:
\begin{equation}
  \footnotesize
  \sum_i \bar{\pi}_{ij} = u_j \left( \sum_i \mathbf{W}[i,j] v_i \right) = \| \bar{\mathbf{q}} \|_1
\end{equation}
\begin{equation}
  \footnotesize
  \sum_j \bar{\pi}_{ij} = \left( u_j \sum_i \mathbf{W}[i,j] \right) v_i = \| \bar{\mathbf{p}} \|_1
\end{equation}
To constraint these two equations simultaneously, one can calculate $v_i$ and $u_j$ by alternately updating the following:
\begin{equation}
  \label{eq:sinkhorn_iteration}
  \footnotesize
  u_j^{t+1} = \frac{\| \bar{\mathbf{q}} \|_1}{\sum_i \mathbf{W}[i, j] v_i^t}, v_i^{t+1} = \frac{ \| \bar{\mathbf{p}} \|_1}{\sum_j \mathbf{W}[i, j] u_j^{t+1}}
\end{equation}
Eqn. \eqref{eq:sinkhorn_iteration} is also known as the Sinkhorn-Knopp Iteration
updating equations. After repeating this iteration $T$ times, the approximate optimal plan $\boldsymbol{\bar{\pi}}^\star$ can be obtained:
\begin{equation} \label{eq:sinkhorn_iteration_final}
  \footnotesize
  \boldsymbol{\bar{\pi}}^\star = \text{diag}(v) \mathbf{W} \text{ diag}(u)
\end{equation}
where $\gamma$ and $T$ are empirically set to 0.1 and 50.

\begin{figure*}[t]
\centering
\includegraphics[width = 0.6\columnwidth]{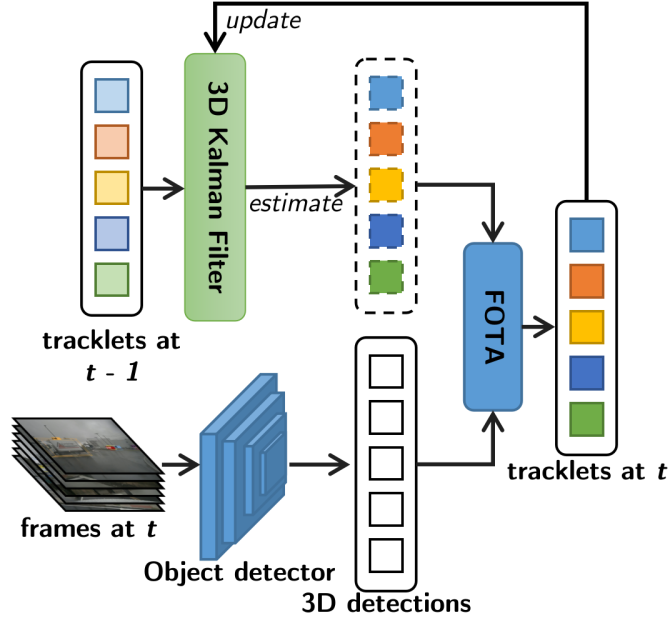}
\caption{The SAGA-Track workflow.}\label{fig:saga_track}
\end{figure*}

In summary, as shown in Fig. \ref{fig:saga_track}, SAGA-Track with a multi-camera matching algorithm is performed in the following steps:
\begin{enumerate}
  \item
        Estimating the next location of the track $\mathbf{tr}_i^{(t-1)}$ using motion model, e.g., Kalman filter.
  \item
        Computing world-coordinate-based distance metrics between $\hat{\mathbf{tr}}_i^{(t)}$ and $\mathbf{o}_{j}^{(t)}$.
  \item
        Solving One-to-many FOTA assignments as in Eqn. \eqref{eq:sinkhorn_iteration_final}. 
  \item
        Updating $i$-th tracklet $\mathbf{tr}_i^{(t)}$ based on assigned objects.
\end{enumerate}

In addition to the proposed track-by-detection scheme for multi-camera, we introduce a novel end-to-end framework including detector, motion model, tracker, and assignment steps in a single model in the next section \ref{sec:ot_e2e}. This end-to-end framework can be fully aware of objects' movement globally rather than taking pre-computed detection as SAGA-Track.

\subsection{End-to-end Learning MC-MOT via FOTA Loss}
\label{sec:ot_e2e}

\begin{figure*}[t]
\centering
\includegraphics[width =\columnwidth]{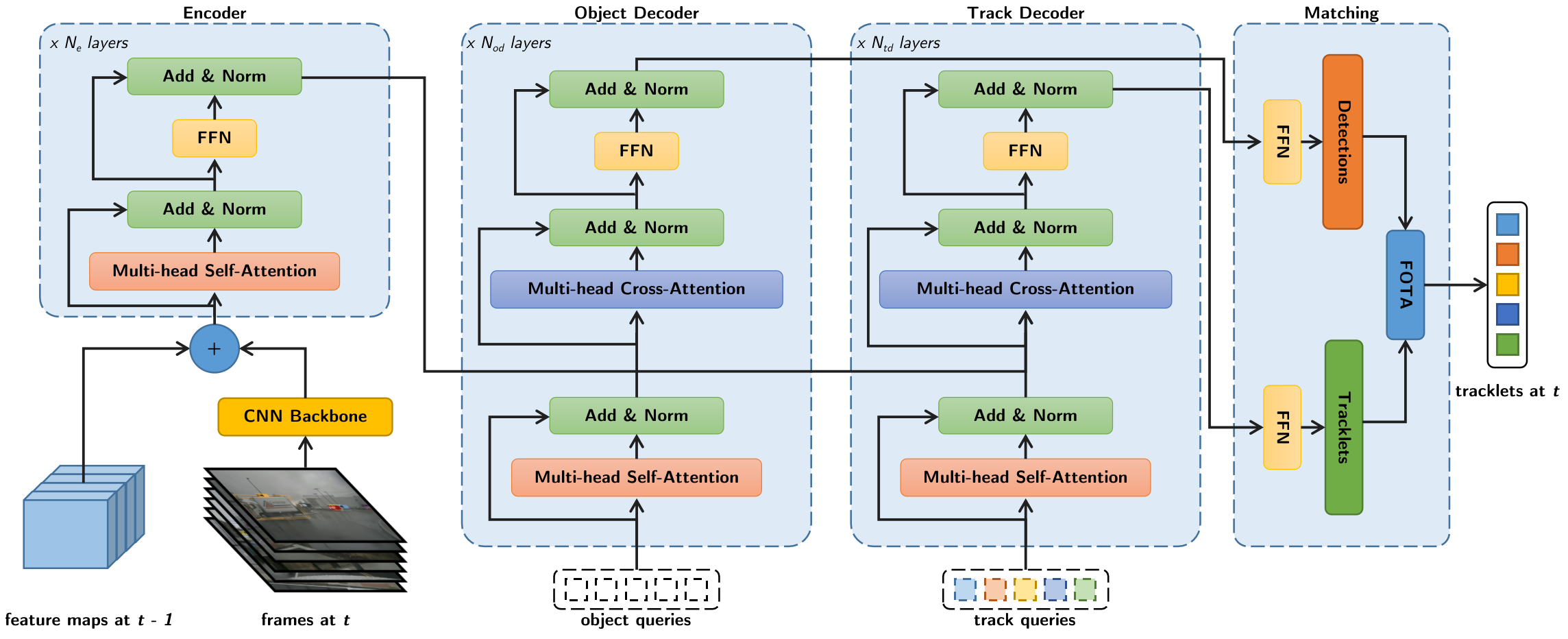}
\caption{The proposed SAGA-TrackNet via Transformer Encoder-Decoder tracking framework}\label{fig:framework}
\end{figure*}

In this section, we further leverage the proposed FOTA into the design of the end-to-end learning network for MC-MOT, named \textbf{SAGA-TrackNet}.

Our proposed architecture consists of an encoder, two decoders, and a box-matching layer. The one-to-many assignment algorithm is implemented to provide the final tracking results from detection and tracked boxes as in Fig. \ref{fig:framework}.

\subsubsection{Model Structure}

The SAGA-TrackNet structure is based on transformer encoder, and decoder tracking frameworks \cite{sun2020transtrack, meinhardt2021trackformer} and contains multi-head attention layers. These layers can be self-attention or cross-attention, i.e., keys and queries are the same or different.

\textbf{Encoder.} Features of the current and previous frame from a camera are extracted by a backbone CNN network, e.g., Resnet-50 \cite{He_2016_CVPR}, and stacked together with other cameras. Features of the previous frame were saved to avoid re-computation. The encoder of SAGA-TrackNet then encodes those feature maps into keys for being used in the following decoders.

\textbf{Object Decoder.}
To detect new objects on each camera, the model takes multiple sets of learnable parameters, named object queries, as a set of objects of interest in the images to match with keys, i.e., the feature maps generated by the encoder, and provides the outputs as "detected boxes."

\textbf{Track Decoder and Matching.}
Simultaneously, the model takes tracked objects in the previous frames as the track query to infer the location of the corresponding tracked objects in the current frame and provides "tracked boxes."
It is performed using the decoder block as it learns object motion similar to the Kalman filter. We can also utilize this Track Decoder block as a motion model to refine any off-the-shelf 3D object detectors by treating the track queries as placeholders and feeding detector predictions to this block. The motion modeling ablative study is further discussed in Subsection \ref{ssec:ablation_study}.
During testing, the matching layer then performs the association of detected objects and tracked objects via FOTA.  
During training, a set prediction loss is computed for all $M + N$ output predictions in two steps: (a) loss for detecting object at frame $t - 1$ using $M$ object queries; (b) loss for tracking objects from (a) and detecting new objects at frame $t$ with all $M$ object queries and $N$ track queries from the frame $t - 1$. 
This prediction loss, computed based on the assignment obtained from FOTA between ground truth and prediction, is described in the following Subsection \ref{sec:training_e2e}.

\subsubsection{Model Training}
\label{sec:training_e2e}

This section presents the procedure for training our proposed end-to-end learning networks. 
\textbf{Training Data.} We train our proposed SAGA-TrackNet on a large-scale dataset, i.e., nuScenes, a training set with 750 scenes of 20s each, and use its validation set for our ablation study. Each training sample contains a chunk size of two consecutive frames from a training sequence.

\textbf{FOTA Loss for Modeling Training.} To compute this loss function, we also need to compute the assignment $\pi_{ij}$ between one of the ground-truth tracks $\mathcal{T}^{\star}_i$ or background to the joint set of object and track query predictions $\hat{\mathbf{o}}_j^{(t)}$. Similar to the OT-based assignment described in Subsection \ref{sec:bipartite_matching}, the assignment is computed based cost matrix using a pre-defined distance between bounding boxes.
Let us denote $G^{(t)} \subset G$ as the subset of ground-truth track ID at time step $t$. Then we assign each detection from step (a) to its corresponding ground-truth track ID $i$ from the set $G^{(t-1)} \subset G$.
These two sets are explicitly assigned to the ground-truth objects in frame $t$ as $G^{(t)} \cap G^{(t-1)}$.
Another set of ground-truth track ID is $G^{(t)} \backslash G^{(t-1)}$, which includes tracks not visible at time $t$. The last set is the new object not yet being tracked ground-truth objects, i.e., new objects, as $G^{(t - 1)} \backslash G^{(t)}$ to be matched with $M$ object queries.

\begin{equation}
  \footnotesize
  \min_{\bar{\boldsymbol{\pi}} \in \Pi(\bar{\mathbf{p}}, \bar{\mathbf{q}})} \langle \bar{\mathbf{C}}, \bar{ \boldsymbol{\pi}} \rangle_F = \underset{\bar{\boldsymbol{\pi}} \in \Pi(\bar{\mathbf{p}}, \bar{\mathbf{q}})}{\min} \overset{N}{\underset{i=1}{\sum}} \overset{M}{\underset{j=1}{\sum}} c_{ij} \pi_{ij}
\end{equation}
Using a similar extension as in Eqn. \eqref{eq:extend_cost}, the cost matrix $\mathbf{C}$ can now be defined as in Eqn. \eqref{eq:CC}.
\begin{equation} \label{eq:CC}
  \footnotesize
  \mathbf{C} = ( c_{ij} ) = -\hat{p}_{\pi_{ij}}(\text{cls}_i) + \cC_{\text{box}} \left(\mathcal{T}_{i}^{\star(t)}, \hat{\mathbf{o}}^{(t)}_{j} \right)
\end{equation}
where $\text{cls}_i$ is the class id of the object and $\cC_{\text{box}}$ term penalizes bounding box differences by a linear combination of a $\ell_1$ distance and a Generalized Intersection over Union \cite{rezatofighi2019generalized} as defined in Eqn. \eqref{eq:g_iou},
\begin{equation}
  \footnotesize
  \cC_{\text{box}} = \lambda_{\ell_1} \| \mathcal{T}_{i}^{\star(t)} - \hat{\mathbf{o}}^{(t)}_{j} \|_1 + \lambda_{GIoU} \cC_{GIoU} \left(\mathcal{T}_{i}^{\star(t)}, \hat{\mathbf{o}}^{(t)}_{j} \right)
\end{equation}
We use set prediction loss to measure the set of predictions for $M$ detection and $N$ tracklets compared with ground-truth tracks in terms of classification and location (bounding boxes).
Set-based loss is based on the optimal bipartite matching (described in Sections \ref{sec:bipartite_matching} and \ref{sec:ot_e2e}) between $M$ detection and ground-truth objects while $N$ tracklets will be matched with boxes from previous frames. The final MC-MOT set prediction loss is defined as in Eqn. \eqref{eq:cbox}.
\begin{equation} \label{eq:cbox}
  \small
  \mathcal{L}_{\text{MC-MOT}} (\mathcal{T}^{\star}, \hat{\mathbf{o}}^{(t)}, \boldsymbol{\pi}) = \overset{M + N}{\underset{j=1}{\sum}}  \mathcal{L}_{\text{query}} (\mathcal{T}^{\star}, \hat{\mathbf{o}}_j^{(t)}, \boldsymbol{\pi})
\end{equation}
The output predictions that do not match any ground-truth tracks will be assigned to the background class $\text{cls}_i = 0$. We indicate the ground-truth track matched with prediction $i$ by $\pi_{ij} = 1$ and define the loss per query as in Eqn. \eqref{eq:lquery}.
\begin{equation} \label{eq:lquery}
\footnotesize  
  \cL_{\text{query}} \left( \mathcal{T}^{\star}, \hat{\mathbf{o}}_j^{(t)}, \boldsymbol{\pi} \right) =
    \begin{cases}
      -\hat{p}_{\pi_{ij}}(\text{cls}_i) +\mathcal{L}_{\text{box}} \left(\mathcal{T}_{i}^{\star(t)}, \hat{\mathbf{o}}^{(t)}_{j} \right) & \text{if } \pi_{ij} = 1 \\
      -\hat{p}_{\pi_{ij}}(0) & \text{if } \pi_{ij} = 0 \\
    \end{cases}
\end{equation}
where
$\mathcal{L}_{box}$ is the combination of the $\ell_1$ loss and the generalized Intersection over Union (IoU) \cite{rezatofighi2019generalized} for 3D boxes.

\subsubsection{Model Inference}
\label{sec:testing_e2e}

During testing, SAGA-TrackNet performs feature encoding, object decoding, and track decoding, then one-to-many matching for two consecutive frames from all cameras. The output features from the backbone network are stored in combination with the subsequent frames. We also keep tracked objects alive and allow them to rebirth to handle occlusion or disappearing quickly.

\section{Experimental Results}

In this section, we detail the benchmark datasets and metrics in Subsection \ref{ssec:data_metrics}. Then, the setups for all experiments and the ablation study will be presented in Subsection \ref{ssec:exp_setup} and \ref{ssec:ablation_study} respectively. The comparisons with the SOTA methods will be detailed in Subsection \ref{ssec:compare_results} on a large-scale Tracking Challenge, i.e. nuScenes Vision Track.

\subsection{Benchmark Datasets and Metrics}
\label{ssec:data_metrics}

\textbf{nuScenes\footnote{License CC BY-NC-SA 4.0} Dataset \cite{caesar2020nuscenes}} is one of the large-scale datasets for Autonomous Driving with 3D object annotations. It contains 1,000 videos of 20-second shots in a setup of 6 cameras, i.e. 3 front and 3 rear ones,  with a total of 1.4M images. It also provides 1.4M manually annotated 3D bounding boxes of 23 object classes based on LiDAR data. This dataset has an official split of 700, 150, and 150 videos for training, validation, and testing, respectively.

The proposed method is evaluated using both detection and tracking metrics described in \cite{caesar2020nuscenes}.

\textbf{Detection Metrics.} A commonly used metric, i.e. \textit{Mean Average Precision (mAP)}, is defined as a match using a 2D center distance on the ground plane instead of intersection over union cost for nuScenes detection challenges.

Similarly, other motion-related metrics are also defined in nuScenes, such as \textit{Average Translation Error (ATE)} measuring Euclidean center distance in 2D in meters, \textit{Average Scale Error (ASE)} computing as $1 - IOU$ after aligning centers and orientation, \textit{Average Orientation Error (AOE)} measuring the smallest yaw angle difference between prediction and ground-truth in radians, \textit{Average Velocity Error (AVE)} measuring the absolute velocity error in $m/s$ and \textit{Average Attribute Error (AAE)} computing $1 - acc$, where $acc$ is the attribute classification accuracy.
We also use the \textit{nuScenes Detection Score (NDS)} that is based on a simple additive weighting of the mean of all metrics above.

\textbf{Tracking Metrics.}  The tracking performance is measured using the popular \textit{CLEAR MOT} metrics \cite{bernardin2008evaluating} including \textit{MOTA}, \textit{MOTP}, ID switch (\textit{IDS}), mostly tracked (\textit{MT}), mostly lost (\textit{ML}), fragmented (\textit{FRAG}). Similar to nuScenes, we use two accumulated metrics introduced in \cite{weng2020ab3dmot} as the main metrics, including the average over the MOTA metric (\textit{Average MOTA (AMOTA)}) and the average over the MOTP metric (\textit{Average MOTP (AMOTP)}).

\subsection{Experiments Setup}
\label{ssec:exp_setup}

\begin{table}[!t]
  \footnotesize
  \centering
  \caption{3D object detectors with and without using our motion model in terms of detection metrics on the nuScenes validation set for Vision Detection challenge. \textbf{MM} - indicates using our SAGA-TrackNet motion decoder}
  \resizebox{1.0\columnwidth}{!}{
  \begin{tabular}{|l|c|c|c|c|c|c|c|c|}
      \hline
      \textbf{Method}                                     & \textbf{MM} & \textbf{mAP} $\uparrow$ & \textbf{NDS} $\uparrow$ & \textbf{mATE} $\downarrow$ & \textbf{mASE} $\downarrow$ & \textbf{mAOE} $\downarrow$ & \textbf{mAVE} $\downarrow$ \\
      \hline
      \multirow{2}{*}{MonoDIS \cite{Simonelli_2019_ICCV}} & \xmark      & 0.2976                  & 0.3685                  & 0.7661                     & 0.2695                     & \textbf{0.5839}            & 1.3619                     \\
      \cline{2-8}
                                                          & \cmark      & \textbf{0.3019}         & \textbf{0.3893}         & \textbf{0.6558}            & \textbf{0.2410}            & 0.6787                     & \textbf{1.3209}            \\
      \hline
      \multirow{2}{*}{CenterNet \cite{zhou2019objects}}   & \xmark      & 0.3027                  & 0.3262                  & 0.7152                     & 0.2635                     & \textbf{0.6158}            & 1.4254                     \\
      \cline{2-8}
                                                          & \cmark      & \textbf{0.3487}         & \textbf{0.4016}         & \textbf{0.5417}            & \textbf{0.2023}            & 0.6317                     & \textbf{1.3094}            \\
      \hline
      \multirow{2}{*}{KM3D \cite{2009.00764}}             & \xmark      & 0.2763                  & 0.3201                  & 0.7495                     & 0.2927                     & 0.4851                     & \textbf{1.4322}            \\
      \cline{2-8}
                                                          & \cmark      & \textbf{0.3503}         & \textbf{0.4117}         & \textbf{0.6998}            & \textbf{0.2323}            & \textbf{0.4661}            & 1.5341                     \\
      \hline
    \end{tabular}}
  \label{tab:nuscene_detection_results}
\end{table}

The proposed SAGA-TrackNet is trained with two consecutive frames where the extracted features in the previous time step $t-1$ are stored and stacked with the features of the current time step to encode object key features to predict the location of new and existing objects at time step $t$.
Then, Mini-batch (chunk of two) gradient descent is employed with an Adam optimizer to learn all the parameters in the attention layers.
All the layers and algorithms are implemented in PyTorch \cite{paszke2019pytorch}, based on Trackformer \cite{meinhardt2021trackformer}, TransTrack \cite{sun2020transtrack} and Deformable DETR \cite{zhu2020deformable}. The best configuration of layers is chosen empirically as three stacking self-attention layers with four heads and three stacking cross-attention layers with 16 heads. With a batch size of 512 chunks, the model converged at about 100 epochs.

\subsection{Ablation Study}
\label{ssec:ablation_study}

In this section, we present some experiments to ablate the effect of each component of the proposed framework.  Particularly, this section aims to demonstrate the following:
\textbf{1.} how this motion modeling can help improve 3D object detectors;
\textbf{2.} better motion modeling with track decoder layers in SAGA-TrackNet;
\textbf{3.} how the combination of the external input and data association method affects the tracking performance. We also compare the processing time of these methods as well as the end-to-end solution.

\begin{table}[!t]
  \centering
  \caption{Motion Errors comparison of different motion models. \textbf{MM} - indicates using our SAGA-TrackNet motion decoder}
    \resizebox{0.7\columnwidth}{!}{
      \begin{tabular}{|c|c|c|c|c|}
        \hline
        \textbf{Method}              & \textbf{mATE} $\downarrow$ & \textbf{mASE} $\downarrow$ & \textbf{mAOE} $\downarrow$ & \textbf{mAVE} $\downarrow$ \\
        \hline
        3D KF \cite{weng2020ab3dmot} & 0.8153                     & 0.5155                     & 0.7382                     & 1.6186                     \\
        LSTM \cite{Chaabane2021deft} & 0.8041                     & 0.4548                     & 0.6744                     & 1.6139                     \\
        \hline
        \textbf{MM}                  & \textbf{0.7132}            & \textbf{0.4388}            & \textbf{0.5677}            & \textbf{1.4189}            \\
        \hline
      \end{tabular}
    }
    \label{tab:motion_errors}
\end{table}

\begin{table*}[!t]
  \centering
  \caption{Comparison of tracking metrics between Kuhn-Munkres (KM) and our proposed FOTA for different types of distance. The computation cost in terms of FPS is measured on both constructing the distance metric and optimizing the cost assignment.}
  \resizebox{1.0\textwidth}{!}{
    {\begin{tabular}{|c|c|c|c|c|c|c|c|c|c|c|c|c|c|c|c|c|c|}
          \hline
          \textbf{Cost matrix}                                       & \textbf{Association} & \textbf{AMOTA} $\uparrow$ & \textbf{AMOTP} $\downarrow$ & \textbf{MOTAR} $\uparrow$ & \textbf{MOTA} $\uparrow$ & \textbf{MOTP} $\downarrow$ & \textbf{MT} $\uparrow$ & \textbf{ML} $\downarrow$ & \textbf{IDS} $\downarrow$ & \textbf{FRAG} $\downarrow$ & \textbf{FPS} $\uparrow$ \\
          \hline
          \multirow{2}{*}{2D GIoU \cite{rezatofighi2019generalized}} & KM                   & 0.071                     & 1.675                       & 0.298                     & 0.065                    & 0.923                      & 327                    & 3,865                    & 6,421                     & 3,254 & \textbf{34.7}                    \\
          \cline{2-12}
                                                                     & \textbf{FOTA}        & \textbf{0.088}            & \textbf{1.604}              & \textbf{0.375}            & \textbf{0.097}           & \textbf{0.898}             & \textbf{356}           & \textbf{3,659}           & \textbf{6,313}            & \textbf{3,124}  & 31.8             \\
          \hline
          \multirow{2}{*}{3D GIoU \cite{rezatofighi2019generalized}} & KM                   & 0.083                     & 1.612                       & 0.423                     & 0.92                     & 0.892                      & 386                    & 3,964                    & 3,560                     & 2,725                & \textbf{32.9}        \\
          \cline{2-12}
                                                                     & \textbf{FOTA}        & \textbf{0.094}            & \textbf{1.563}              & \textbf{0.445}            & \textbf{0.102}           & \textbf{0.883}             & \textbf{481}           & \textbf{3,988}           & \textbf{2,250}            & \textbf{2,219}  & 29.2             \\
          \hline
          \multirow{2}{*}{Mahalanobis \cite{chiu2020probabilistic}}  & KM                   & 0.143                     & \textbf{1.473}              & 0.501                     & 0.143                    & 0.869                      & 678                    & 3,676           & 1,998                     & 1,867     & \textbf{25.6}                   \\
          \cline{2-12}
                                                                     & \textbf{FOTA}        & \textbf{0.242}            & 1.541                       & \textbf{0.551}            & \textbf{0.234}           & \textbf{0.823}             & \textbf{1,419}         & \textbf{2,980}                    & \textbf{522}              & \textbf{1,590}  & 21.3             \\
          \hline
        \end{tabular}}
  }
  \label{tab:association_ablation_study}
\end{table*}

\textbf{Improving 3D Object Detector.}
Table \ref{tab:nuscene_detection_results} demonstrates that the combination of baselines object detector and our motion model (i.e. the Track Decoder) achieves better results than the original detector. In this experiment, we initialize detected objects at previous frames as inputs to the track queries. The Track Decoder takes that set of objects and then combines it with frame features produced by the Encoder to refine the location of pseudo-"tracked boxes". The best result is achieved with the combination of KM3D object detector \cite{2009.00764} and our motion model since it is guided by decoded locations from our transformation procedure as described in Section \ref{sec:ot_e2e}.

\textbf{The Role of Motion Model.}
Motion models are particularly essential in \textit{dynamic} MC-MOT settings since cameras are moving with the vehicle. In this experiment, we evaluate the effectiveness of different motion modeling methods on detection performance. We use the locations predicted by motion models to compare with ground-truth locations in terms of motion-related metrics. In such a way, we can evaluate how well the motion model captures and predicts the motion of tracked objects.  We compare with two other commonly used motion models, i.e. 3D Kalman Filter \cite{weng2020ab3dmot} and LSTM \cite{Chaabane2021deft}.
As shown in Table \ref{tab:motion_errors}, our SAGA-TrackNet gives better results than a classical object state prediction technique, i.e. 3D Kalman Filter used in \cite{weng2020ab3dmot} and a deep learning-based technique, i.e. LSTM module, used in \cite{Chaabane2021deft}.

\begin{table}[!t]
    \centering
    \caption{Comparison of 3D tracking performance of different tracklet-detection matching approaches}
    \resizebox{0.8\columnwidth}{!}{
      \begin{tabular}{|c|c|c|c|c|c|c|c|c|c|c|}
        \hline
        \textbf{Inputs}                              & \textbf{Assoc.}                       & \textbf{MOTA} $\uparrow$ & \textbf{MOTP} $\downarrow$ & \textbf{IDS} $\downarrow$ & \textbf{FRAG} $\downarrow$ \\	
        \hline
        MOT                                          & Re-ID \cite{Qian_2020_CVPR_Workshops} & 0.197                    & 0.838                      & 1,691                     & 2,036           \\
        \hline
        DET                                          & AB3DMOT \cite{weng2020ab3dmot}        & 0.164                    & 0.853                      & 1,608                     & 1,733               \\ 
        \hline
        DET                                          & \textbf{FOTA}                         & 0.234                    & \textbf{0.823}             & \textbf{522}              & \textbf{1,590}                    \\
        \hline
        \multicolumn{2}{|c|}{\textbf{SAGA-TrackNet}} & \textbf{0.237}                        & 0.833                    & 732                        & 2,000                                  \\
        \hline
      \end{tabular}
    }
    \label{tab:track_ablation_study}
\end{table}

\textbf{Comparison of Different Distance Cost and Matching Algorithms.}
The proposed assignment module operates on a global cost matrix, which is computed from a detection set and a track set by different types of distance, i.e. Mahalanobis distance \cite{chiu2020probabilistic} as defined in Eqn. \eqref{eq:mahalanobis}, Bird's Eye View 2D, and 3D bounding box GIoU \cite{rezatofighi2019generalized} as defined in Eqn. \eqref{eq:g_iou} between the estimated object states and the detected object bounding boxes.
Then, the Sinkhorn iterative is employed as described in Section \ref{sec:bipartite_matching} with the maximum number of iterations being 100. Compared to Kuhn-Munkres (KM) algorithm, our framework inherits the merit of one-to-many matching and yields better results on assignment metrics and tracking metrics with a slight increase in computation cost in distance matrices construction and optimization (shown in Table \ref{tab:association_ablation_study}).
The performance of our proposed FOTA algorithm is also better than other tracklet-detection matching methods as shown in Table \ref{tab:track_ablation_study}.

\begin{figure*}[!t]
  \centering
  \includegraphics[width=1.0\linewidth]{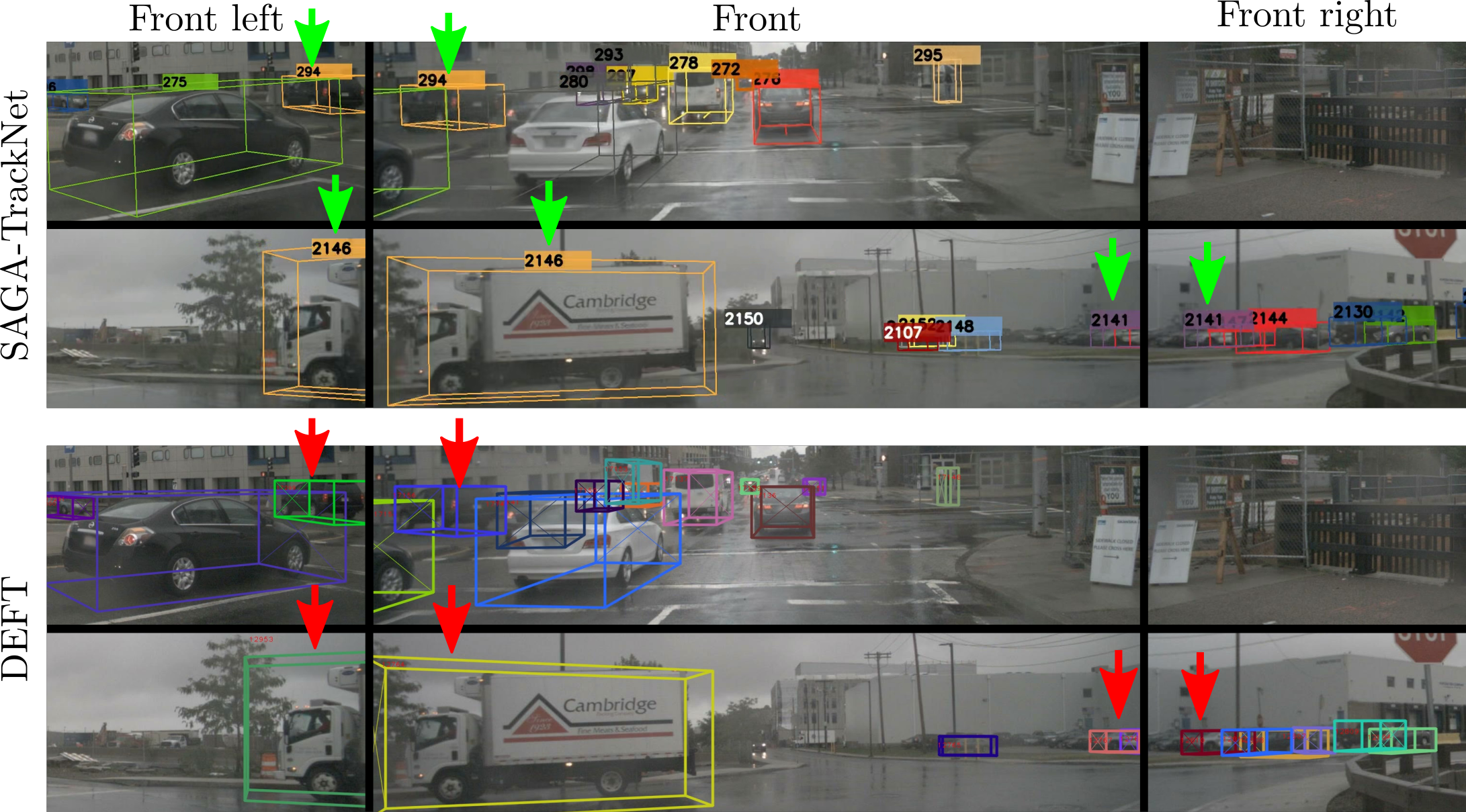}
  \caption{Our proposed MC-MOT method (top) can recognize a positive tracking case compared with an SC-MOT system that has no object correlation linking module across all cameras (bottom). Green arrows indicate true positive tracking samples, red arrows indicate false negative tracking samples. Best viewed in color and zoom in.}
  \label{fig:compare_tracking}
\end{figure*}

\begin{table*}[!t]
  \centering
  \caption{Comparison of 3D tracking performance on the nuScenes \textbf{validation set} for Vision Track challenge.}
  \resizebox{\textwidth}{!}{
    \begin{tabular}{|l|c|c|c|c|c|c|c|c|c|c|c|c|}
      \hline
      \textbf{Method}                                                     & \textbf{Glo. Assoc.} & \textbf{AMOTA} & \textbf{AMOTP} & \textbf{MOTAR} & \textbf{MOTA} $\uparrow$ & \textbf{MOTP} $\downarrow$ & \textbf{RECALL} $\uparrow$ & \textbf{MT} $\uparrow$ & \textbf{ML} $\downarrow$ & \textbf{IDS} $\downarrow$ & \textbf{FRAG} $\downarrow$ \\	
      \hline
      MonoDIS \cite{Simonelli_2019_ICCV} + AB3DMOT \cite{weng2020ab3dmot} & \xmark               & 0.045          & 1.793          & 0.202          & 0.047                    & 0.927                      & 0.293                      & 395                    & 3,961                    & 6,872                     & 3,229                      \\ 
      CenterTrack \cite{zhou2020tracking}                                 & \xmark               & 0.068          & 1.543          & 0.349          & 0.061                    & \textbf{0.778}             & 0.222                      & 524                    & 4,378                    & 2,673                     & 1,882                      \\ 
      DEFT \cite{Chaabane2021deft}                                        & \xmark               & 0.213          & 1.532          & 0.49           & 0.183                    & 0.805                      & \textbf{0.4}               & 1,591                  & 2,552                    & 5,560                     & 2,721                      \\ 
      QD-3DT \cite{Hu2021QD3DT}                                           & \xmark               & 0.242          & 1.518          & 0.58           & 0.218                    & 0.81                       & 0.399                      & \textbf{1,600}         & \textbf{2,307}           & 5,646                     & 2,592                      \\
      \hline
      MonoDIS \cite{Simonelli_2019_ICCV} + AB3DMOT \cite{weng2020ab3dmot} & \cmark (baseline)    & 0.027          & 1.959          & 0.263          & 0.045                    & 1.010                      & 0.049                      & 169                    & 5,304                    & 1,903                     & 2,947                      \\ 
      CenterTrack \cite{zhou2020tracking}                                 & \cmark (baseline)    & 0.056          & 1.578          & 0.478          & 0.102                    & 0.782                      & 0.201                      & 454                    & 4,784                    & 1,173                     & 1,682                      \\ 
      DEFT \cite{Chaabane2021deft}                                        & \cmark (baseline)    & 0.185          & 1.638          & 0.601          & 0.193                    & 0.81                       & 0.32                       & 1,019                  & 3212                     & 1,793                     & 1,647                      \\ 
      QD-3DT \cite{Hu2021QD3DT}                                           & \cmark (baseline)    & 0.237          & 1.544          & 0.564          & 0.226                    & 0.826                      & 0.375                      & 1,414                  & 3,007                    & 1,593                     & 1,623                      \\
      \hline
      \textbf{SAGA-Track (Ours)}                                          & \cmark (FOTA)        & 0.242          & 1.541          & 0.551          & 0.234                    & 0.823                      & 0.375                      & 1,419                  & 2,980                    & \textbf{522}              & \textbf{1,590}             \\
      \textbf{SAGA-TrackNet (Ours)}                                       & \cmark (End-to-end)  & \textbf{0.261} & \textbf{1.485} & \textbf{0.626} & \textbf{0.237}           & 0.833                      & \textbf{0.4}               & 1,302                  & 2,978                    & 732                       & 2,000                      \\

      \hline
    \end{tabular}
  }
  \label{tab:nuscene_val_track_results}
\end{table*}

\begin{table*}[!t]
  \centering
  \caption{Comparison of 3D tracking performance on the nuScenes \textbf{test set} for Vision Track challenge.}
  \resizebox{\textwidth}{!}{
    \begin{tabular}{|l|c|c|c|c|c|c|c|c|c|c|c|c|}
      \hline
      \textbf{Method}                     & \textbf{Glo. Assoc.} & \textbf{AMOTA} & \textbf{AMOTP} & \textbf{MOTAR} & \textbf{MOTA} $\uparrow$ & \textbf{MOTP} $\downarrow$ & \textbf{RECALL} $\uparrow$ & \textbf{MT} $\uparrow$ & \textbf{ML} $\downarrow$ & \textbf{IDS} $\downarrow$ & \textbf{FRAG} $\downarrow$ \\	
      \hline
      CenterTrack \cite{zhou2020tracking} & \xmark               & 0.046          & 1.543          & 0.231          & 0.043                    & \textbf{0.753}             & 0.233                      & 573                    & 5,235                    & 3,807                     & 2,645                      \\ 
      DEFT \cite{Chaabane2021deft}        & \xmark               & 0.177          & 1.564          & 0.484          & 0.156                    & 0.770                      & 0.338                      & \textbf{1,951}         & 3,232                    & 6,901                     & 3,420                      \\ 
      QD-3DT \cite{Hu2021QD3DT}           & \xmark               & 0.217          & 1.550          & 0.563          & 0.198                    & 0.773                      & \textbf{0.375}             & 1,893                  & \textbf{2,970}           & 6,856                     & 3,001                      \\
      \hline
      \textbf{SAGA-Track}                 & \cmark (FOTA)        & 0.226          & 1.574          & 0.616          & \textbf{0.218}           & 0.791                      & 0.317                      & 1,540                  & 3,825                    & \textbf{797}              & \textbf{1,953}             \\
      \textbf{SAGA-TrackNet}              & \cmark (End-to-end)  & \textbf{0.242} & \textbf{1.480} & \textbf{0.627} & 0.209                    & 0.756                      & 0.340                      & 1,469                  & 4,148                    & 870                       & 2,153                      \\

      \hline
    \end{tabular}
  }
  \label{tab:nuscene_test_track_results}
\end{table*}

\subsection{Comparison against State-of-the-Art Methods}
\label{ssec:compare_results}

In this section, we compare our proposed framework with other vision-based (without using LiDAR or RADAR information) tracking approaches, which are the top in nuScenes vision only tracking challenge leaderboard.

\textbf{Comparison against Tracking Methods on Validation set.}
This experiment compares our proposed method with other vision-based methods, including QD-3DT \cite{Hu2021QD3DT}, MonoDIS \cite{Simonelli_2019_ICCV} + AB3DMOT \cite{weng2020ab3dmot}, CenterTrack \cite{zhou2020tracking}, and DEFT \cite{Chaabane2021deft} which are the tops of nuScenes vision-only tracking challenge. As we can see in Table \ref{tab:nuscene_val_track_results}, we outperform the top approach, i.e. QD-3DT, in most of the metrics. Fig. \ref{fig:compare_tracking} illustrates the key factor that helps improve the tracking performance: we perform appearance matching across cameras in addition to motion modeling. It shows that our proposed method (top) can assign object ID globally between cameras compared with DEFT \cite{Chaabane2021deft} (bottom).  
Our method beats the SOTA method, i.e. QD-3DT \cite{Hu2021QD3DT} on most of the main metrics, such as AMOTA, AMOTP, MOTAR, MOTA, Recall, IDSwitch, and FRAG, which are related to how well our method groups tracklet IDs and regresses object's bounding boxes. For fair comparison and to preserve the originality and uniqueness of those methods, such as the LSTM motion model of DEFT \cite{Chaabane2021deft}, the offset head of CenterTrack \cite{zhou2020tracking}, we implement a simple global association as the baseline, which takes MOT output results from those approaches and then adopts several empirical rules and heuristics to determine and filter out duplicated objects, including IOU thresholding and box merging as similar to ELECTRICITY \cite{Qian_2020_CVPR_Workshops}.

\textbf{Comparison against Tracking Methods on Test set.}
We submitted our result on the official competition platform EvalAI\footnote{ \href{https://eval.ai/web/challenges/challenge-page/476/leaderboard/1321\#leaderboardrank-52}{https://eval.ai/web/challenges/challenge-page/476/leaderboard/1321}}. As it can be referred to the tracking challenge leaderboard on Vision track at nuScenes' homepage\footnote{\href{https://www.nuscenes.org/tracking/?externalData=no\&mapData=no\&modalities=Camera}{https://www.nuscenes.org/tracking/}} and Table \ref{tab:nuscene_test_track_results}, our method performs better than QD-3DT \cite{Hu2021QD3DT} and DEFT \cite{Chaabane2021deft} significantly on IDS (870 vs. 6,856 and 6,901) and slightly on AMOTA (0.242 vs. 0.217 and 0.177), this behavior is similar to the validation results in Table \ref{tab:nuscene_val_track_results}.

\section{Conclusions}

This paper has introduced a new global association approach to solving the dynamic MC-MOT problem for AV. The proposed framework can learn how to perform tracking frame-by-frame in an end-to-end manner given frames from multi-camera to extract features, encode object key features, decode new objects' locations, decode tracked objects' locations, and global association tracklets with detection. These tasks are enhanced with self-attention and cross-attention layers to capture structures and motion across cameras. The experiments have shown performance improvements up to 6.4\% and a decrease in IDSwitch error from 3,807 to 870 in a large-scale AV dataset regarding vision-based detection and tracking accuracy.

\newpage

\bibliographystyle{model2-names}
\bibliography{mybibfile}

\clearpage
\clearpage

\end{document}